\pdfoutput=1
\documentclass[11pt]{article}

\usepackage{acl}
\usepackage{times}
\usepackage{latexsym}

\usepackage[T1]{fontenc}
\usepackage[utf8]{inputenc}

\usepackage{microtype}

\usepackage{inconsolata}

\usepackage{graphicx}
\usepackage[utf8]{inputenc} % allow utf-8 input
\usepackage[T1]{fontenc}    % use 8-bit T1 fonts
\usepackage{hyperref}       % hyperlinks
\usepackage{url}            % simple URL typesetting
\usepackage{booktabs}       % professional-quality tables
\usepackage{amsfonts}       % blackboard math symbols
\usepackage{nicefrac}       % compact symbols for 1/2, etc.
\usepackage{microtype}      % microtypography
\usepackage{xcolor}         % colors
\usepackage{graphicx} % Required for inserting images  
\usepackage[export]{adjustbox}
\usepackage{multirow}
\usepackage{graphicx} 
\usepackage{array}
\usepackage{longtable}
\usepackage{amsmath}
\usepackage{caption}
\usepackage{mathtools}
\usepackage{float}
\usepackage{placeins}
\usepackage{microtype}
\title{From Knowledge to Treatment: Large Language Model Assisted Biomedical Concept Representation for Drug Repurposing}

\author{Chengrui Xiang\textsuperscript{*} \and Tengfei Ma\textsuperscript{*} \and Xiangzheng Fu \\ \and \bf{Yiping Liu} \and \bf{Bosheng Song} \and \bf{Xiangxiang Zeng}\textsuperscript{†} \\ % <- 删除这行的 \\
  College of Computer Science and Electronic Engineering, Hunan University \\
  Changsha, Hunan Province, China \\
  \texttt{\{mmtd,tfma,xzeng\}@hnu.edu.cn}
}

\begin{document}
\maketitle
\renewcommand{\thefootnote}{\fnsymbol{footnote}}
\footnotetext[1]{*Equal contribution \quad †Corresponding author}
\begin{abstract}
Drug repurposing plays a critical role in accelerating treatment discovery, especially for complex and rare diseases. Biomedical knowledge graphs (KGs), which encode rich clinical associations, have been widely adopted to support this task. However, existing methods largely overlook common-sense biomedical concept knowledge in real-world labs, such as mechanistic priors indicating that certain drugs are fundamentally incompatible with specific treatments. To address this gap, we propose LLaDR, a Large Language Model-assisted framework for Drug Repurposing, which improves the representation of biomedical concepts within KGs. Specifically, we extract semantically enriched treatment-related textual representations of biomedical entities from large language models (LLMs) and use them to fine-tune knowledge graph embedding (KGE) models. By injecting treatment-relevant knowledge into KGE, LLaDR largely improves the representation of biomedical concepts, enhancing semantic understanding of under-studied or complex indications. Experiments based on benchmarks demonstrate that LLaDR achieves state-of-the-art performance across different scenarios, with case studies on Alzheimer’s disease further confirming its robustness and effectiveness. 
Code is available at https://github.com/xiaomingaaa/LLaDR.
% Drug repurposing is a critical task in drug discovery, aiming to identify new therapeutic uses for existing drugs for complex diseases. Existing methods  Clinical-centered knowledge  While existing approaches have shown promising results, they primarily focus on discovering new therapeutic uses using a data-driven paradiagram, often overlooking 
% While most existing approaches rely heavily on data-driven paradigms, they often overlook the general biomedical knowledge embedded in entities. To address this limitation, we propose integrating large language models (LLMs) into the Drug Repurposing Knowledge Graph (DRKG) through a novel framework called Drug Repurposing Fine-Tuning (DR-FIT). DR-FIT leverages large language models (LLMs) to generate semantically rich discriptions of biomedical entities and integrates this information into the knowledge graph through a fine-tuning process. By incorporating both LLM-derived textual knowledge and the structural information of DRKG, DR-FIT enhances the expressiveness of entity representations. Extensive experiments on a real-world drug repurposing task demonstrate that DR-FIT achieves significant improvements over structure-based, GNN-based, and language model-based models in identifying effective drug-disease associations. These results highlight the effectiveness of injecting general biomedical knowledge via LLMs to substantially enhance the expressiveness and informativeness of DRKG embeddings.
\end{abstract}

\section{Introduction}
Drug repurposing has emerged as an effective strategy to accelerate drug development by identifying new therapeutic uses for existing drugs~\cite{pushpakom2019drug,huang2024foundation,inoue2024drugagent}. With the increasing complexity of biological systems and the growing availability of heterogeneous biomedical data~\cite{chen2024trialbench}, there is a critical need for computational approaches that can efficiently integrate and reason over these large-scale sources~\cite{wei2024drugrealign}.
\begin{figure} % 40%
  \centering % 
  \includegraphics[width=1\columnwidth]{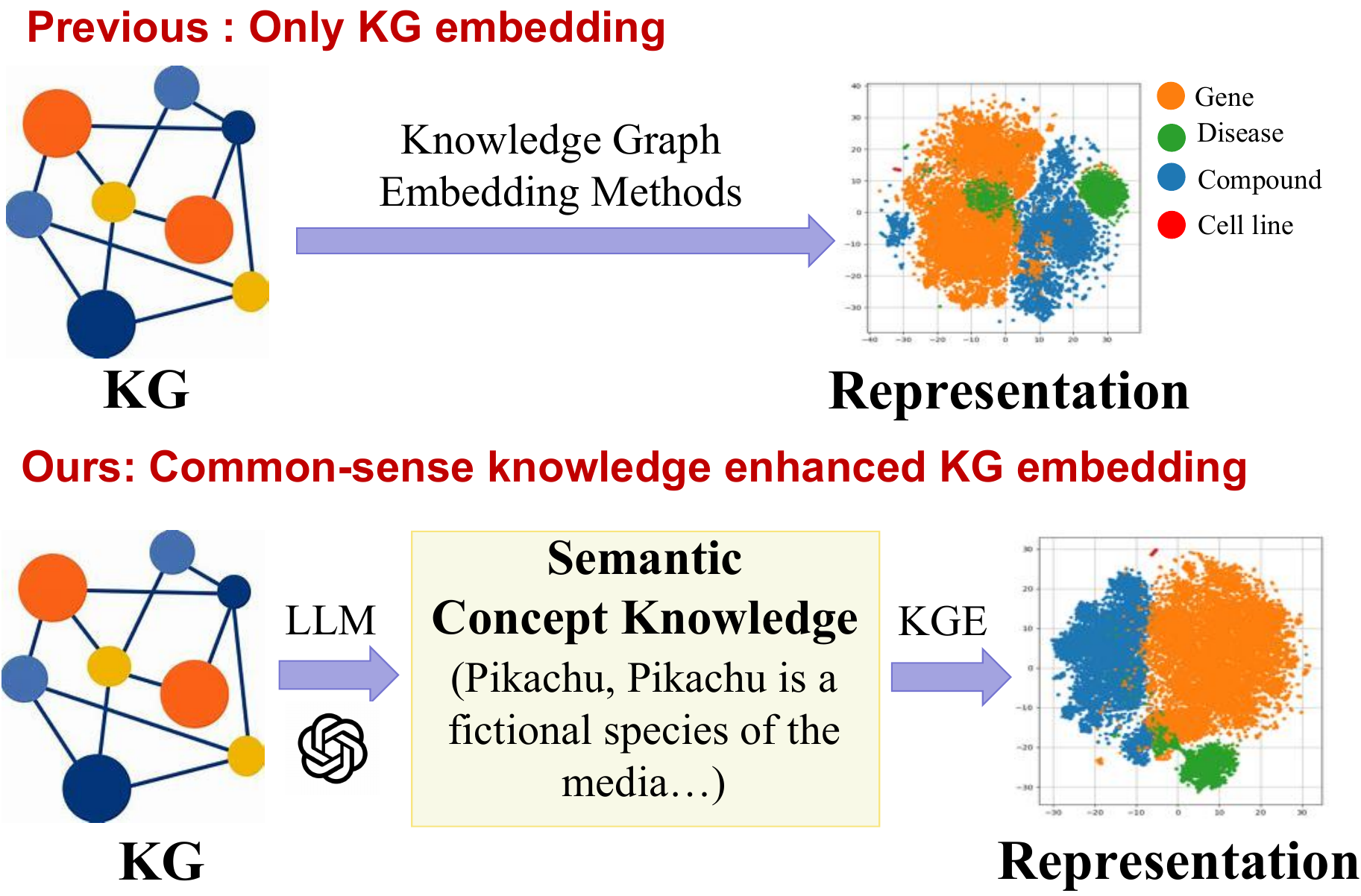} % 
  \caption{Comparison of standard KG embedding (top) and LLaDR (bottom). LLaDR incorporating semantic concept knowledge generates more  meaningful representations, leading to better separation of entities.} %
  \label{fig:intro1}
\end{figure}
Traditional drug repurposing relies heavily on expert-driven analysis of medical literature and clinical data, requiring interdisciplinary collaboration across pharmacology, chemistry, and medicine~\cite{samborskyi2017strategic}. This process is time-consuming and resource-intensive, often leading to low throughput~\cite{hodos2016silico}. Recent advances in deep learning have improved drug repurposing performance by learning complex representations of drugs and targets from data~\cite{zhao2022geometric,su2022predicting,zhao2023fusing}, enhancing both accuracy and interpretability~\cite{lee-lee-2024-repurformer}. However, these models often overlook structured domain knowledge from clinical research and biomedical ontologies~\cite{tayebi2024ekgdr}. To address this, several works~\cite{bang2023biomedical,tayebi2024ekgdr,huang2024foundation} have incorporated knowledge graph embedding (KGE) techniques to model biomedical entities and relations, enabling more interpretable and knowledge-aware drug repurposing. These methods leverage structured biomedical knowledge to improve prediction and support reasoning over complex biological interactions.

However, an important yet underexplored aspect of biomedical KGs is the presence of general-purpose biomedical concept knowledge—common-sense mechanistic constraints that are critical for safe and effective treatment decisions. For example, acetylcholinesterase inhibitors (e.g., Donepezil) are effective for Alzheimer’s disease but contraindicated in Parkinson’s disease psychosis due to potential cholinergic overstimulation~\cite{goldman2014treatment}. Such concept-level priors are rarely captured by structure-only KG embedding models, leading to limited reliability when generalizing to rare or mechanistically complex diseases. 

To address this gap, we propose LLaDR (Large Language Model-assisted Drug Repurposing), a framework that integrates concept-level semantics extracted from LLMs into KG embeddings, promoting better representations of biomedical concepts (e.g., compound, disease, gene, and cell lines in Figure~\ref{fig:intro1}). LLaDR generates enriched biomedical concept representations by prompting LLMs with textual discriptions and using these embeddings to guide KGE fine-tuning. By injecting treatment-aware knowledge into KGs, LLaDR enhances both predictive accuracy and robustness in drug repurposing tasks. 
In summary, our main contributions are as follows:
(1) We propose the first framework that integrates common-sense biomedical concept knowledge to enhance KG-based drug repurposing;
(2) We introduce a novel KG fine-tuning approach that leverages LLM-derived concept knowledge to improve the semantic expressiveness of knowledge graph embeddings;
(3) Extensive experiments show that LLaDR achieves state-of-the-art performance on standard benchmarks and exhibits strong robustness under KG noise and semantic perturbations.

\section{Related Work}

\subsection{Drug Repurposing}
Recent advances in drug repurposing leverage graph learning and deep representation models to capture complex drug-target-disease relationships~\cite{zhao2022geometric,su2022predicting,zhao2023fusing}. Transformer-based architectures have further improved molecular generation by modeling repurposing-aware chemical semantics~\cite{lee-lee-2024-repurformer}.
To enhance interpretability and knowledge grounding, several methods incorporate biomedical knowledge graphs, encoding structured relations between entities for more reliable predictions~\cite{tayebi2024ekgdr,bang2023biomedical}. These approaches extend beyond data-driven correlations by integrating curated ontologies and multi-relational graphs.
Foundation models have recently emerged as powerful tools for repurposing, combining large-scale clinical data and language model reasoning to support explainable, clinician-aligned decisions~\cite{huang2024foundation,inoue2024drugagent}.
However, most existing approaches focus on structural or statistical associations, often neglecting high-level biomedical constraints such as contraindications and mechanistic incompatibilities—factors essential for clinically valid repurposing. To address this, we propose a concept-aware framework that explicitly incorporates biomedical priors into the reasoning process, enabling more trustworthy and mechanistically grounded repurposing decisions.

\subsection{LLM incorporated Knowledge Representation}
 Recent advances in integrating pre-trained language models (PLMs) with knowledge graphs (KGs) have shown promise in aligning unstructured text with structured knowledge. KG-BERT~\cite{yao2019kg}, PKGC~\cite{lv2022pre}, and KG-LLM~\cite{yao2025exploring} adopt classification or prompt-based training to inject KG information into PLMs. FAE~\cite{verga2020facts} and KEPLER~\cite{wang2021kepler} further combine cross-modal fusion with graph-enhanced contrastive objectives to bridge textual and relational signals. However, these methods are computationally intensive due to the need to sample or enumerate large numbers of triples.
To improve efficiency, StAR~\cite{wang2021structure} and CSProm-KG~\cite{chen2023dipping} fuse graph and PLM embeddings, but are limited to small-scale models and overlook the structured semantic priors encoded by LLMs. Fully prompt-based approaches~\cite{wei2024kicgpt,bi2024lpnl} offer stronger generalization but are expensive and less suited to specialized tasks like drug repurposing.
In contrast, LLaDR leverages LLMs to extract biomedical concept semantics and integrates them into KG embedding through a lightweight, hierarchical alignment framework, enabling scalable and concept-aware reasoning for drug repurposing.

\section{LLaDR Framework}
\label{others}
\begin{figure*}[ht]
  \centering
  \includegraphics[width=1.0\linewidth]{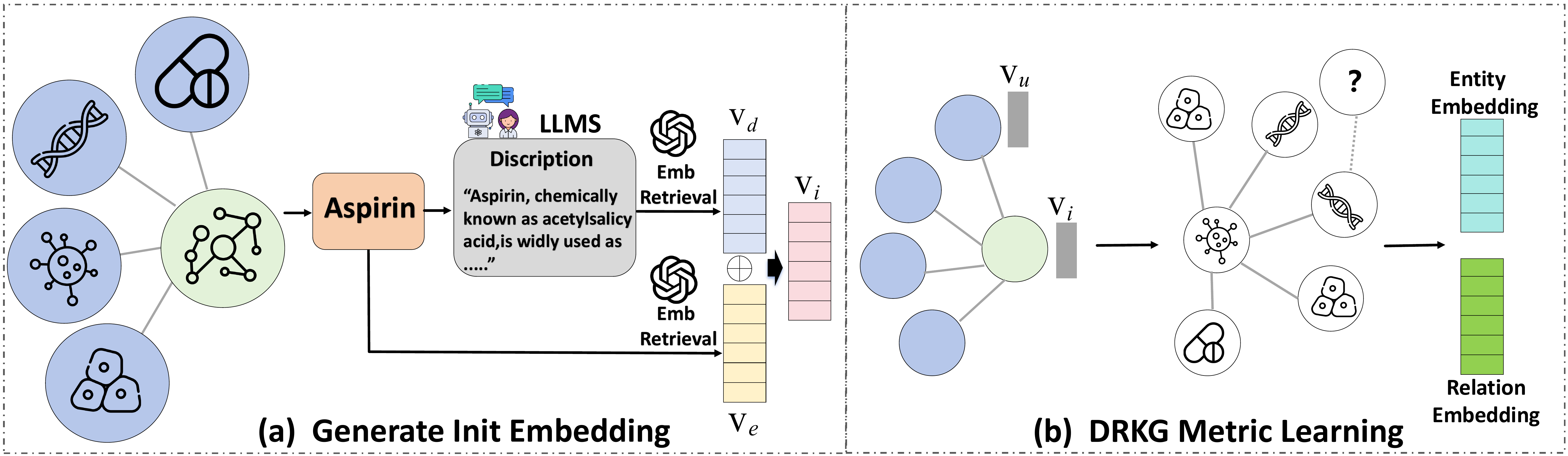}
  \caption{Overview of LLaDR. Input and Output are highlighted at each step. Step 1: Obtain text embeddings for all entities in DRKG, achieved by merging word embeddings with discription embeddings retrieved from
LLMs. Step 2: We utilize the initial vectors generated by LLMs and apply them to the training of knowledge graphs, thereby obtaining the final embeddings of entities and relations enriched with comprehensive semantics.}
  \label{fig:intro}
\end{figure*}

We present LLaDR (as shown in Fig.\ref{fig:intro}), a novel framework specifically designed for fine-tuning knowledge graph embeddings through the strategic integration of textual information derived from open knowledge sources. This comprehensive framework consists of two core operational phases that work in tandem: 
(1) \textbf{Generating discription :} During this initial stage, we employ advanced large language models (LLMs) such as GPT-4 to systematically generate rich, context-aware textual discriptions for every entity within the knowledge graph. These AI-generated discriptions are subsequently processed through state-of-the-art sentence embedding techniques like Sentence-BERT to create high-dimensional vector representations, which are then stored alongside the original KG embeddings in parallel npy format files for entity-discription alignment. 
(2) \textbf{Knowledge Graph Fine-Tuning:} This crucial stage implements a multi-layered optimization approach where the initial DRKG embeddings undergo enhancement through three synergistic mechanisms - firstly by concatenating textual embeddings through weighted averaging, secondly by applying geometric alignment constraints to preserve structural relationships, and thirdly by incorporating semantic consistency regularization terms that bridge the gap between symbolic KG representations and neural text embeddings. The iterative fine-tuning process leverages adaptive learning rate scheduling and contrastive loss functions to ultimately produce knowledge representations that simultaneously capture topological accuracy from the graph structure and nuanced semantic depth from open-domain textual knowledge. 
\FloatBarrier

\subsection{Generating Discription}

We input entities into a large language model to obtain their semantically rich discriptions, utilizing the model's sophisticated natural language processing capabilities to generate detailed, nuanced, and contextually grounded representations of these entities. By leveraging the model's advanced understanding of language and its ability to synthesize information from vast datasets, we ensure that the resulting discriptions capture not only the explicit attributes of the entities but also their implicit relationships, connotations, and broader contextual significance. This process guarantees that the discriptions are comprehensive, encompassing both the surface-level features and the deeper semantic layers of the entities, while also maintaining high levels of accuracy and relevance to their specific domains or applications. The integration of the large language model's outputs thus provides a robust foundation for downstream tasks, enabling more precise and meaningful interactions with the entities in question.

\noindent\textbf{Step 1: Entity Embedding Initialization}  serves as the foundational step in this procedure, where we are provided with a comprehensive set of entities denoted as $E = \{e_1, \dots, e_{|E|}\}$ within a Knowledge Graph (KG). The primary objective of this step is to enhance the semantic representations of these entities by leveraging the capabilities of a Large Language Model (LLM) to generate detailed and informative discriptions for each entity.

To achieve this, for every individual entity $e_i$ within the set $E$, we utilize a carefully designed prompt template to query the LLM. An example of such a template could be phrased as follows: "Briefly describe [entity] in a concise yet informative manner, adhering to the format '[entity] is a [discription]'." This structured approach ensures that the generated discriptions are not only consistent in format but also rich in semantic content, thereby facilitating the subsequent steps in the embedding process.

By systematically iterating through each entity in the set $E$
and applying this method, we aim to construct a robust and semantically enriched representation of the entire entity collection, which will serve as the basis for further processing and analysis within the Knowledge Graph framework.

Subsequently, the entity embedding $v_e^i \in \mathbb{R}^{\text{dim}(f)}$ and discription embedding $v_d^i = f(d_i) \in \mathbb{R}^{\text{dim}(f)}$ are obtained using an embedding model $f$ and concatenated to form the enriched representation:

\begin{equation}
v_i = [v_e^i; v_d^i]. \label{eq:your_label}
\end{equation}

\subsection{Knowledge Graph Fine-Tuning}

\textbf{Step 2: LLaDR} fine-tunes the knowledge graph embeddings by incorporating the text embeddings, which serve as an additional source of information to enrich the structural representations of entities and relations. This integration is guided by two primary constraints designed to optimize the embeddings for both alignment and predictive accuracy. The first constraint, known as the text embedding deviation constraint, ensures that the structural embeddings derived from the knowledge graph remain closely aligned with their corresponding textual representations, thereby maintaining consistency between the two modalities. The second constraint focuses on the link prediction objective, which directly optimizes the embeddings to accurately predict the relationships between entities within the knowledge graph. By combining these constraints, LLaDR not only enhances the alignment between structural and textual representations but also improves the overall performance and coherence of the knowledge graph, making it more robust and reliable for downstream tasks. This dual-objective approach ensures that the embeddings capture both the semantic nuances from the text and the relational dynamics from the graph structure, resulting in a more  effective representation.

\noindent\textbf{Semantic Anchoring Constraint:} To maintain the original semantic integrity of the embeddings, we implement the semantic anchoring constraint, which is formulated as:

\begin{equation}
L_{\text{anc}} = - \sum_{e_i \in E} d(e_i, v_i') \label{eq:anchor_loss},
\end{equation}
Given that $E$ represents the collection of entities, $e_i$ idenotes the fine-tuned embedding associated with entity $e_i$, $v_i'$ stands for the sliced text embedding of entity $e_i$, and $d(\cdot, \cdot)$ indicates the distance function.

This constraint is critically important for large clusters, where the inherent diversity of entities may lead the fine-tuned embeddings to significantly deviate from their original semantic meanings, potentially undermining the model's ability to generalize. The constraint also plays a vital role when working with sparse knowledge graphs, as it effectively prevents the model from overfitting to the limited and often incomplete structural information that is available. By functioning as a powerful regularization term, it not only mitigates the risk of overfitting but also substantially enhances the overall robustness and reliability of the learned embeddings, ensuring they remain semantically coherent and practically useful across diverse scenarios.

\noindent\textbf{Score Function-Based Fine-Tuning:} LLaDR is a highly versatile and general framework that can be seamlessly integrated with a wide range of existing Knowledge Graph Embedding (KGE) models, as demonstrated in previous studies \cite{daza2021inductive,carvalho2023knowledge}. These KGE models are specifically engineered to learn compact, low-dimensional vector representations of both entities and relations within a knowledge graph, with the primary objective of effectively capturing and preserving the rich semantic and structural information that is inherently embedded within the graph itself. In the context of our research, we place particular emphasis on the task of link prediction, which serves as a critical mechanism to enhance and refine the model's overall capability to make precise and accurate predictions about the potential relationships that may exist between various entities within the knowledge graph. The link prediction loss function associated with this framework is carefully defined and formulated to ensure optimal performance in this regard and decomposed into two components: the positive sample loss $\mathcal{L}_{\text{pos}}$ and the negative sample loss $\mathcal{L}_{\text{neg}}$. $\mathcal{L}_{\text{pos}}$ measures the score of the observed triple $(e_i, r, e_j)$:
    \begin{equation}
    \mathcal{L}_{\text{pos}} = \log \sigma(\gamma - f_r(e_i, e_j)),
    \label{eq:pos_loss}
    \end{equation}
    
\noindent $\mathcal{L}_{\text{neg}}$ computes a penalty term by averaging the scores of negative samples $e_j'$, where these samples are randomly selected from the predefined negative sampling set $N_j$ corresponding to each observed positive triple:
    \begin{equation}
    \mathcal{L}_{\text{neg}} = - \frac{1}{|N_j|} \sum_{e_j' \in N_j} \log \sigma(f_r(e_i, e_j') - \gamma).
    \label{eq:neg_loss}
    \end{equation}
The link prediction loss function, which serves as the core optimization objective in knowledge graph embedding, can be formally expressed as:
\begin{equation}
L_{\text{link}} = - \sum_{(e_i, r, e_j) \in D} \left[ \mathcal{L}_{\text{pos}} + \mathcal{L}_{\text{neg}} \right],
\label{eq:link_loss}
\end{equation}

\noindent here $D$ denotes the complete set of all triples contained within the knowledge graph, while $\sigma$ represents the sigmoid activation function,which is employed to normalize the output values. The scoring function $f_r(\cdot, \cdot)$ , which is explicitly defined by the selected Knowledge Graph Embedding (KGE) model, serves as a critical component for quantifying the compatibility or alignment between the embedding vector of the head entity $e_i$ and the embedding vector of the tail entity $e_j$ , given the specific relation $r$.Additionally $N_j$ corresponds to the set of negative tail entities that are systematically sampled for the positive triple $(e_i, r, e_j)$, where each $e_j'$ signifies the embedding vector of a negative tail entity $e_j'$ within this sampled set. Finally, $\gamma$ is a predefined margin hyperparameter that plays a pivotal role in regulating the training process and ensuring the discriminative power of the learned embeddings.

The link prediction-based fine-tuning process is fundamentally designed to achieve two complementary objectives: first, it systematically minimizes the scoring function $f_r(e_i, e_j)$ for the true, positive triples $(e_i, r, e_j)$, ensuring that these triples are assigned the highest possible compatibility scores by the model. Simultaneously, it rigorously maximizes the margin or separation between the scores of these true triples and the scores of the artificially generated negative triples $(e_i, r, e_j')$ where $e_j'$ represents a corrupted or incorrect tail entity. This dual optimization strategy serves as a powerful inductive bias, compelling the model to distinctly differentiate between valid and invalid relationships by assigning significantly higher scores to the positive triples and substantially lower scores to the negative triples. By enforcing this discriminative scoring behavior, the fine-tuning process effectively imbues the learned embeddings with the rich, localized semantic patterns and relational structures that are inherently present within the knowledge graph, thereby enhancing their ability to capture and preserve the nuanced contextual information that defines the relationships between entities.The margin hyperparameter $\gamma$ further refines this process by controlling the degree of separation required between the scores of positive and negative triples, ensuring that the embeddings not only achieve high discriminative power but also maintain robustness and generalizability across diverse scenarios.

\noindent\textbf{Training Objective:} The training objective of LLaDR is designed to optimize the model by integrating two distinct constraints, each contributing to the overall learning process. The objective function is formulated as follows:

\begin{equation}
L = \zeta_1 L_{\text{anc}} + \zeta_2 L_{\text{link}},
\label{eq:total_loss}  % 用于交叉引用
\end{equation}

\noindent here, $\zeta_1$ and $\zeta_2$ are hyperparameters that determine the relative importance of each constraint in the total loss. The first term, $L_{\text{anc}}$,represents the anchor loss, which ensures that the embeddings of similar entities or relations are pulled closer together in the vector space. The second term,$L_{\text{ink}}$,corresponds to the link loss, which enforces the structural consistency of the knowledge graph by preserving the relationships between entities.

By combining these two constraints, LLaDR achieves a balanced optimization that not only captures the semantic similarities between entities but also maintains the integrity of the relational structure. The hyperparameters $\zeta_1$ and $\zeta_2$ allow for fine-tuning the model's behavior, enabling it to prioritize either the anchor or link constraint based on the specific requirements of the task or dataset. This flexibility makes the model adaptable to various scenarios, ensuring robust performance across different applications.
\section{Experimental Setup}
\label{subsec:experiment_setup}

\textbf{Datasets}. We consider datasets that encompass various domains and sizes, ensuring comprehensive evaluation of the proposed model. Specifically, we utilize the DRKG dataset \cite{ioannidis2022drkg}, a biomedical knowledge graph that includes drugs, diseases, genes, and cell lines, comprising 68,471 entities, 101 relationships, and 4,359,327 triples, which are partitioned into training (3,889,539 triples), validation (429,959 triples), and test (39,829 triples) sets to facilitate robust evaluation while preserving the integrity of the biomedical relationships. Table \ref{tab:dataset_stats} (in Appendix) shows the statistics of DRKG \cite{ioannidis2022drkg}.

\noindent\textbf{Drug Repurposing Task}. In this study, we evaluate the performance of our knowledge graph drug repurposing model using a tail entity replacement strategy. For each test triple,we randomly sample 50 candidate tail entities from a predefined list of disease entities to construct candidate triples. Meanwhile, we selected three therapeutically relevant relation types from DRKG as the relation types for the candidate triples. Due to the randomness introduced by sampling candidate tail entities, each experiment was repeated five times, and the average result was reported.

\noindent\textbf{Metrics}. Following previous works, we use Mean Rank (MR), Mean Reciprocal Rank (MRR), Hits@N (H@N) and Area Under the Curve(AUC) to evaluate link prediction. MR measures the average rank of true entities, lower the better. MRR averages the reciprocal ranks of true entities, providing a normalized measure less sensitive to outliers. Hits@N measures the proportion of true entities in the top N predictions. AUC measures the quality of the ranking.

\noindent\textbf{Baselines}. To benchmark the performance of our proposed model, we compared it against the state of-the-art PLM-based methods including CSProm-KG \cite{chen2022knowledge} and KGT5 \cite{saxena2022sequence}, GNN-based methods including GraphSAGE \cite{tayebi2024ekgdr}, GAT, and Structure-based methods including TransE \cite{carvalho2023knowledge}, DistMult \cite{jiang2023graphcare} and RotatE \cite{dettmers2018convolutional}. In Structure-based methods,we use dgl-ke \cite{zheng2020dgl} as the baseline.

\noindent\textbf{Experimental Strategy}. For PLM-based models, we reproduce CSProm-KG \cite{chen2022knowledge} and KGT5 \cite{saxena2022sequence} for DRKG,for Structure-based KGE models, we assess and present
their best performance using optimal settings, for GNN-based models, we use GraphSAGE \cite{tayebi2024ekgdr} and GAT frameworks for testing. For LLaDR, we use OpenAI’s GPT-3.5-turbo and GPT-4o-mini as the LLM for entity discription generation. Text-embedding-3-small is used for entity embedding initialization.

\subsection{Experimental Results}
\label{subsec:results}

\textbf{Drug Repurposing Performance}. We conducted a drug repurposing experiment to rigorously evaluate LLaDR, with the results meticulously summarized in Table \ref{result_main}. In this experiment, we thoroughly assessed the accuracy of LLaDR, The experimental outcomes clearly show LLaDR’s superior performance compared to existing approaches in every evaluation criterion. This advantage is consistently observed throughout all testing scenarios and measurement standards, proving the method’s capability to effectively integrate language model knowledge with graph embedding techniques. The comparative results across different model architectures unequivocally establish LLaDR’s leading position in knowledge graph representation learning.

\begin{table}[h!]
\centering
\resizebox{0.48\textwidth}{!}{%
\begin{tabular}{c|cccccccc}
\hline
\textbf{Backbone} & \textbf{Variants} & \textbf{MR↓} & \textbf{H@10↑} & \textbf{AUC↑}\\ \hline
\multirow{2}{*}{GNN} & GraphSAGE & 6.66 & .813 & .844 \\
                                        & GAT   & 6.64 & .812 & .839  \\
                                        \hline
\multirow{2}{*}{PLM} & KGT5 & 10.48 & .650 & .806 \\
                                        & CSProm-KG & 6.64 & .812 & .839 \\
                                        \hline
\multirow{5}{*}{TransE} & Base &6.28 & .840   & .835 \\
                                        & LLaDR-gpt-3.5   &\textbf{5.39}& \textbf{.871} & \textbf{.850}\\
                                        & LLaDR-gpt-4o-mini   & 5.46 & .864 & .849 \\ & LLaDR-LLama   & 5.40 & .867 & .850 \\
                                        & Best Imprv   & ↑16.5\% & ↑3.6\% & ↑1.7\%  \\
                                        \hline
\multirow{5}{*}{DistMult} & Base &7.55 & .786   & .817 \\
                                        & LLaDR-gpt-3.5   &\textbf{6.06}& \textbf{.839} & \textbf{.843}\\
                                        & LLaDR-gpt-4o-mini   &  6.42 & .821 & .836 \\& LLaDR-LLama   &6.32 & .826 &.838 \\
                                        & Best Imprv   & ↑24.5\% & ↑6.7\% & ↑3.1\% \\
                                        \hline
\multirow{5}{*}{RotatE} & Base &6.06& .850   & .837 \\
                                        & LLaDR-gpt-3.5   &\textbf{5.33}& \textbf{.873} & \textbf{.849}\\
                                        & LLaDR-gpt-4o-mini   &  5.52 & .862 & .844 \\& LLaDR-LLama   &5.49& .862 &.844 \\
                                        & Best Imprv   &↑13.6\% & ↑2.7\% & ↑1.4\% \\
                                        \hline
\hline
\end{tabular}%
}
\caption{The performance of LLaDR and baselines for drug repurposing. The \textbf{bold} denotes the best results.}
\label{result_main}
\end{table}

 As quantitatively demonstrated in Table \ref{result_main}, the LLaDR framework maintains consistent performance advantages across all evaluation metrics when benchmarked against contemporary methodologies on DRKG datasets, including PLM-enhanced architectures, graph neural network implementations, and conventional structure-driven models. This empirical validation underscores LLaDR's capability to effectively harness the semantic comprehension capabilities of large language models for optimizing knowledge graph embedding spaces. Specifically, when employing GPT-3.5-turbo-generated textual discriptions as contextual inputs, the $\mathrm{LLaDR}_{\mathrm{DistMult}}$ variant achieves significant performance differentials compared to baseline systems: a 24.5\% relative improvement in Mean Rank (MR) metric for entity alignment tasks and an 6.7\% enhancement in HITS@10 (H@10). Furthermore, the framework demonstrates robust discriminative capacity with a 3.1\% increase in Area Under Curve (AUC) measurements, collectively illustrating its effectiveness in preserving structural integrity while integrating linguistic patterns from LLMs across multiple evaluation dimensions. The performance improvements are consistently observed under different experimental configurations and data sampling conditions, confirming the reliability of these quantitative findings. The comparative results clearly establish LLaDR's superiority in both structural and semantic aspects of knowledge graph representation learning. We show more results in Appendix~\ref{appdx:main_result}
\vspace{0.5cm} 

\noindent\textbf{Prompt}.To validate the impact of different prompts on the quality of generated text, we designed three prompts for generating entity discriptions: a no-prompt, an original prompt, and a refined prompt. The specific three prompts and full experiment results are provided in Appendix~\ref{appdx:prompt}. The large language model used in this experiment is GPT-4o-mini.
 \begin{table}[h!]
\centering
\resizebox{0.48\textwidth}{!}{%
\begin{tabular}{c|cccccccc}
\hline
\textbf{Backbone} & \textbf{Variants} & \textbf{MR↓} & \textbf{H@10↑} & \textbf{AUC↑}\\ \hline
\multirow{3}{*}{TransE} & noprompt & 5.64 & .856 & .840 \\
                                        & original prompt   & 5.46  & .864 & .849  \\ & good prompt   & 5.43 & .867 & .850  \\
                                        \hline
\multirow{3}{*}{DistMult} & noprompt & 7.05 & .800 & .822 \\
                                        & original prompt   & 6.42 & .821 & .836  \\ & good prompt   & 6.34 & .824 & .838  \\
                                        \hline
\multirow{3}{*}{RotatE} & noprompt & 5.82 & .851 & .832 \\
                                        & original prompt   &5.52  &.862  & .844  \\ & good prompt   &5.51 & .863 & .844 \\
                                        \hline
 \hline
\end{tabular}%
}
\caption{LLaDR's performance on different prompts for drug repurposing. }
\label{result_prompt}
\end{table}

\noindent\textbf{Temperature}.To validate the impact of temperature on the text generation quality of large language models, we carefully designed a low-temperature experiment, reducing the temperature to 0.1 for text generation, which was used for downstream tasks in drug repositioning. We compared this with the experimental data at the original temperature of 0.7.The full experiment results are provided in Appendix~\ref{appdx:prompt}.
 \begin{table}[h!]
\centering
\resizebox{0.48\textwidth}{!}{%
\begin{tabular}{c|cccccccc}
\hline
\textbf{Backbone} & \textbf{Temperature} & \textbf{MR↓} & \textbf{H@10↑} & \textbf{AUC↑}\\ \hline
\multirow{2}{*}{TransE} & low(0.1) & 5.75 & .856 & .845 \\
                                        & original(0.7)   & 5.46  & .864 & .849  \\ 
                                        \hline
\multirow{2}{*}{DistMult} & low(0.1) & 6.44 & .821 & .836 \\
                                        & original(0.7)   & 6.42 & .821 & .836  \\ 
                                        \hline
\multirow{2}{*}{RotatE} & low(0.1) &5.71 & .857 & .840 \\
                                        & original(0.7)   &5.52  &.862  & .844  \\ 
                                        \hline
 \hline
\end{tabular}%
}
\caption{LLaDR's performance on low temperature for drug repurposing. }
\label{result_temp}
\end{table}

\begin{figure*} % 40%
  \centering % 
  \includegraphics[width=1.0\textwidth]{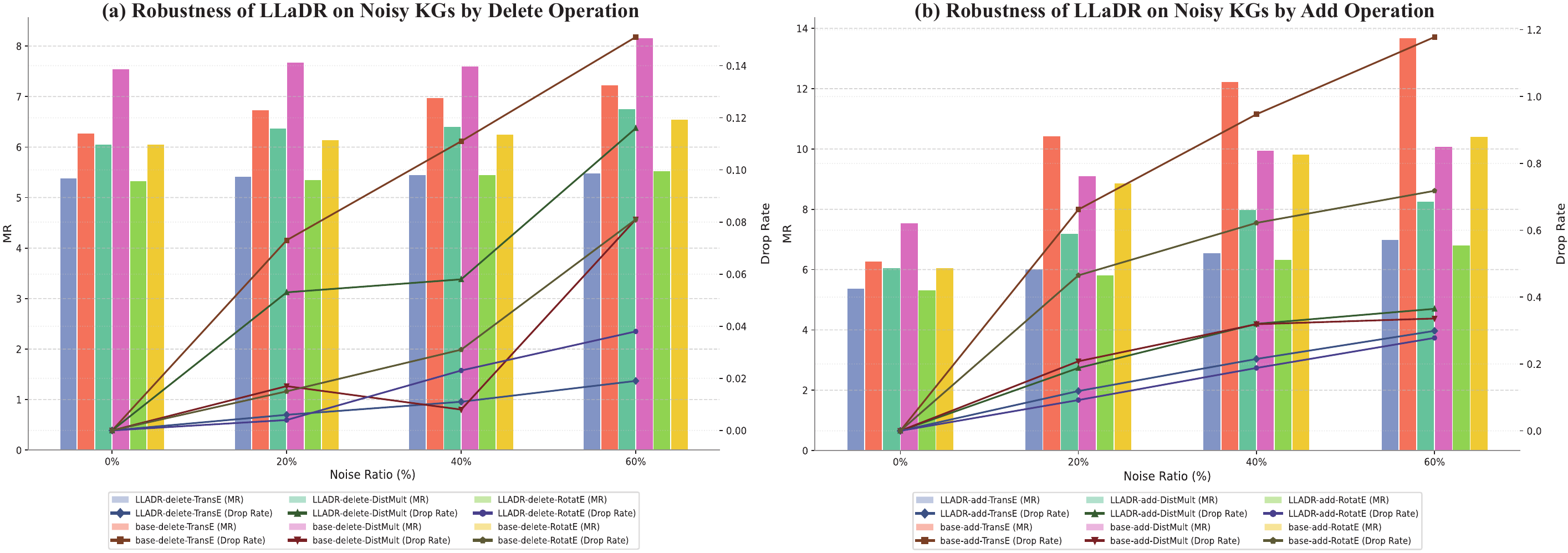} % 
  \caption{Robustness of LLaDR on noisy KGs by deleting or adding unknown associations.} %
  \label{fig:robutness}
\end{figure*}

\noindent\textbf{Robustness}. We conducted a comprehensive robustness testing experiment to rigorously evaluate the robustness of LLaDR. In this experiment, we thoroughly assessed the robustness of LLaDR, as clearly outlined in the paper, and systematically compared its performance to baseline models under varying levels of noise and data perturbation (as shown in Fig.\ref{fig:robutness}). The primary goal was to critically evaluate the models' ability to consistently maintain predictive accuracy in less-than-ideal conditions, ensuring a fair and detailed comparison across all tested scenarios. Remarkably, LLaDR demonstrated superior robustness, maintaining stable performance even under significant noise, primarily due to its integrated regularization mechanisms and the high-quality init embeddings generated by LLMs, which significantly contributed to its resilience. In stark contrast, baseline models \cite{zheng2020dgl}, which lacked such sophisticated mechanisms, exhibited a noticeable and progressive decline in accuracy as noise levels increased, highlighting the limitations of their design. More detailed results are reported in Appendix~\ref{appdx:robust}.

\noindent\textbf{Validity of Knowledge}.We conducted a semantic ablation experiment to validate the effectiveness of discriptions generated by LLMs, analyzing the impact of masking on model performance. We masked 20\%, 40\%, and 60\% of the discriptions generated by the language model, then reused them for drug repositioning tasks, measuring the resulting changes in accuracy and key metrics. Based on the data in Table \ref{mask result}, it can be observed that with the increase in the proportion of semantic masks, the metrics across models decline significantly, confirming the critical role and effectiveness of the discriptions generated by LLMs. The downward trend in performance underscores the importance of preserving the semantic integrity of the discriptions. We provide the full results, including breakdowns and additional analyses, in Appendix~\ref{appdx:mask}, offering a comprehensive view of the experiment's outcomes.

\begin{table}[H]
\centering
\resizebox{0.48\textwidth}{!}{%
\begin{tabular}{c|cccccccc}
\hline
\textbf{Backbone} & \textbf{Variants} & \textbf{MR↓} & \textbf{H@10↑} & \textbf{AUC↑}\\ \hline
\multirow{4}{*}{TransE} & original &5.46 & .864   & .849 \\
                                        & mask 20\%   & 5.96& .839 & .843\\
                                        &mask 40\%   & 6.04 & .838 & .838 \\
                                        & mask 60\%   & 6.12 & .835 & .836  \\
                                        \hline
\multirow{4}{*}{DistMult} & original &6.42 & .821  & .836 \\
                                        & mask 20\%   & 6.46& .820 & .829\\
                                        &mask 40\%   & 6.66 & .813 & .827 \\
                                        & mask 60\%   & 6.74 & .809 & .828  \\
                                        \hline
\multirow{4}{*}{RotatE} &  original &5.52  & .862   & .844 \\
                                        & mask 20\%   & 5.86& .844 & .840\\
                                        &mask 40\%   & 5.93  & .842 & .838 \\
                                        & mask 60\%   & 6.03 & .839 & .836  \\
                                        \hline
\hline
\end{tabular}%
}
\caption{Robustness of LLaDR based on masked semantic discription.}
\label{mask result}
\end{table}

\subsection{Case Analysis}
\begin{figure}[H] % 40%
  \centering % 
  \includegraphics[width=1\columnwidth]{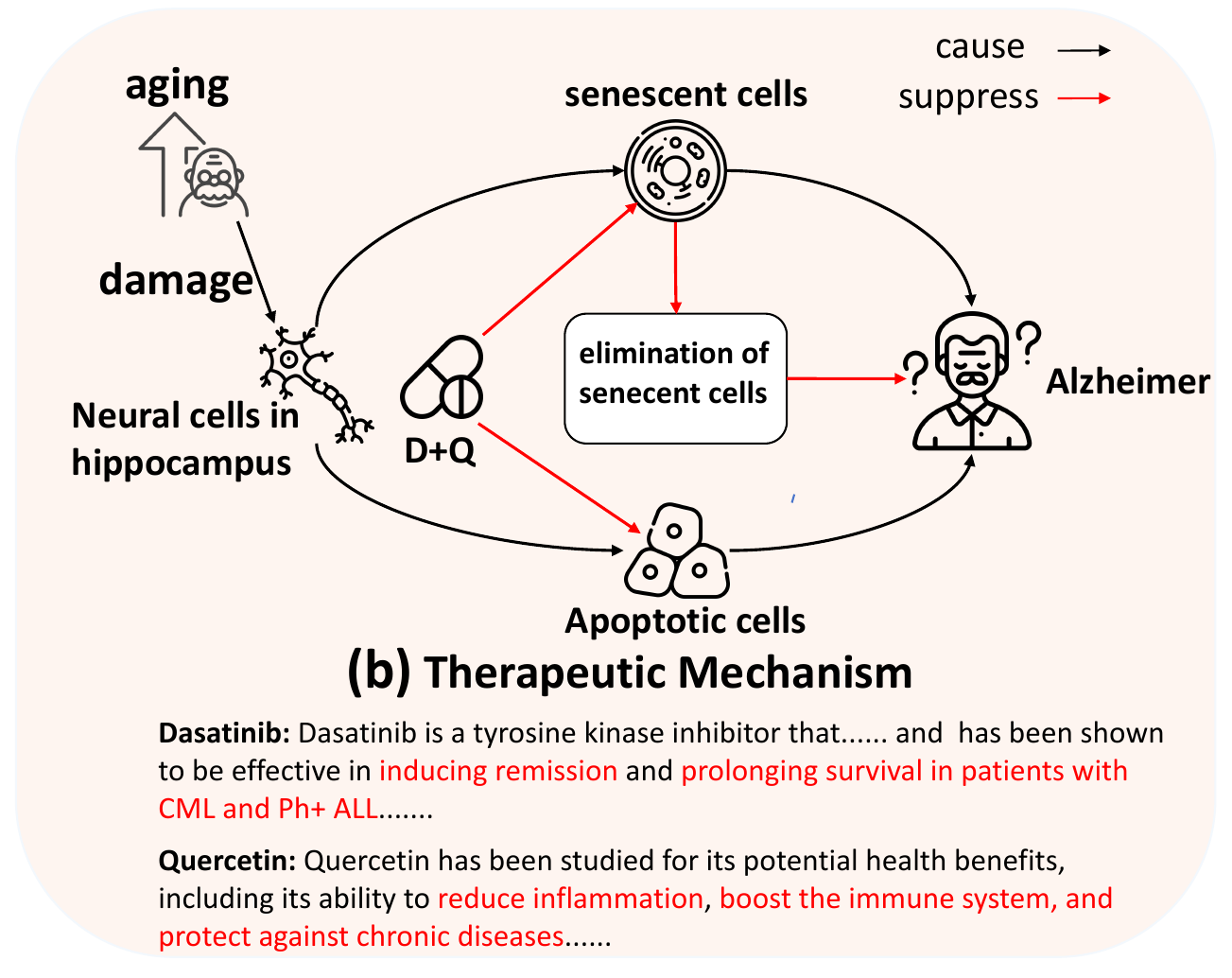} % 
  \caption{The mechanism of action of Dasatinib and Quercetin in Alzheimer's disease.} %
  \label{fig:case}
\end{figure}

In our study, we employed a novel and systematic drug repositioning strategy to identify promising and potentially effective candidates for Alzheimer’s disease treatment. This carefully designed strategy was specifically developed to enhance both the precision and breadth of therapeutic discovery by integrating robust computational frameworks with extensive biomedical data. LLaDR demonstrated a significant and measurable improvement in predictive accuracy, clearly and consistently outperforming a broad range of traditional and widely used methods, including several baseline and benchmark algorithms across various evaluation metrics. Through comprehensive and large-scale screening of its top-ranked compounds across a broad and diverse pharmacological space, we successfully pinpointed several potential therapeutics, including both Dasatinib and Quercetin, which consistently showed strong and reproducible therapeutic potential in the context of Alzheimer’s disease. We further examined the textual discriptions and related biomedical literature of Dasatinib and Quercetin in meticulous detail, and carefully annotated the specific sections that may directly contribute to their high ranking as potential treatments for Alzheimer’s disease (as shown in Fig.\ref{fig:case}). Studies such as \cite{krzystyniak2022combination} provide additional and supporting evidence for the use of Dasatinib and Quercetin in Alzheimer’s therapy, thereby reinforcing their clinical relevance and practical utility. These findings collectively and robustly highlight LLaDR's superior capability in identifying clinically relevant drug candidates with significant translational promise and therapeutic potential. Appendix\ref{appdx:drug} lists the top 10 predictions, further validating, confirming, and strengthening the robustness, consistency, and overall reliability of our proposed approach.

\section{Conclusion}
This study introduces LLaDR, a framework enhancing the Drug Repurposing Knowledge Graph (DRKG) by combining structural embeddings with LLM-derived semantics. It surpasses traditional methods in drug-disease prediction accuracy by jointly leveraging topological and contextual information.

The approach captures structural and semantic relationships critical for drug discovery, applicable to biomedical KGs like disease-gene networks. Future work may extend to larger graphs and multimodal data.

In summary, LLaDR enhances DRKG for drug repositioning, providing a scalable solution for knowledge graph improvements in drug discovery.

\section*{Limitations}

While LLaDR significantly enhances drug repurposing by integrating LLM semantics with knowledge graph embeddings and demonstrates remarkable robustness to noise, certain limitations still persist and require careful consideration and further investigation in the future. Key challenges include scalability issues, domain generalization difficulties, and ethical concerns, which must be addressed for broader adoption and practical implementation in real-world settings. (1) \textbf{Dependence on LLM:} Errors or biases in LLM-generated discriptions (e.g., GPT-3.5-turbo, GPT-4o-mini) can significantly affect performance and reliability, potentially limiting its effectiveness and practical utility over time. (2) \textbf{Scalability Challenges:} Computational costs may rise substantially for larger graphs (millions of entities), posing practical barriers to widespread use and deployment at scale. (3) \textbf{Interpretability Concerns:} The "black-box" nature of LLMs complicates clinical adoption and raises transparency issues, limiting trust and acceptance among key stakeholders. Addressing these challenges could unlock LLaDR’s potential for clinical decision support and wider use in practice.

\section*{Acknowledgments}
This work was supported by the National Natural Science Foundation of China (2023ZD0120902 to X.Z.; U22A2037 to X.Z.; 62425204 to X.Z.; 62122025 to X.Z.; 62450002 to X.Z.; and 62432011 to X.Z.). This work was supported by the Beijing Natural Science Foundation (L248013  to X.Z.)

% This document has been adapted
% by Steven Bethard, Ryan Cotterell and Rui Yan
% from the instructions for earlier ACL and NAACL proceedings, including those for
% ACL 2019 by Douwe Kiela and Ivan Vuli\'{c},
% NAACL 2019 by Stephanie Lukin and Alla Roskovskaya,
% ACL 2018 by Shay Cohen, Kevin Gimpel, and Wei Lu,
% NAACL 2018 by Margaret Mitchell and Stephanie Lukin,
% Bib\TeX{} suggestions for (NA)ACL 2017/2018 from Jason Eisner,
% ACL 2017 by Dan Gildea and Min-Yen Kan,
% NAACL 2017 by Margaret Mitchell,
% ACL 2012 by Maggie Li and Michael White,
% ACL 2010 by Jing-Shin Chang and Philipp Koehn,
% ACL 2008 by Johanna D. Moore, Simone Teufel, James Allan, and Sadaoki Furui,
% ACL 2005 by Hwee Tou Ng and Kemal Oflazer,
% ACL 2002 by Eugene Charniak and Dekang Lin,
% and earlier ACL and EACL formats written by several people, including
% John Chen, Henry S. Thompson and Donald Walker.
% Additional elements were taken from the formatting instructions of the \emph{International Joint Conference on Artificial Intelligence} and the \emph{Conference on Computer Vision and Pattern Recognition}.

\bibliography{custom}
\appendix
\onecolumn
\onecolumn
\section{Appendix}
\subsection{Datasets}
\label{appdx:dataset}
\begin{table}[H]
  \centering
  \scalebox{0.75}{ % 缩放系数 0.7-0.9 之间调整
  \begin{tabular}{lccccc}
    \hline
    \textbf{Dataset} & \textbf{Entities} & \textbf{Relations} & \textbf{Train} & \textbf{Valid} & \textbf{Test} \\ 
    DRKG & 68,471 & 101 & 3,889,539 & 429,959 & 39,829 \\
    \hline
  \end{tabular}
  }
  \caption{We have meticulously divided the comprehensive Drug Repurposing Knowledge Graph (DRKG) into three distinct parts: 3,889,539 entries for the training set, 429,959 entries for the validation set, and 39,829 entries for the test set. The test set primarily consists of meaningful triplets that describe the intricate relationships between various drugs and diseases, which are specifically used for conducting essential drug repositioning experiments aimed at identifying new therapeutic applications.} % 标题放在表格内容之后
  \label{tab:dataset_stats}
\end{table}

\subsection{Hardware}
\label{appdx:Hardware}
Our experiment uses the RTX 3080 graphics card for training and testing, with a total of 100,000 steps for training. For DRKG, one training session takes 2 hours.

\subsection{Drug Repurposing Results}
\label{appdx:main_result}
{\footnotesize
\begin{longtable}{@{}llc|rrrrr@{}}
\label{tab:model_performance}\\
\toprule
\multicolumn{8}{c}{\textbf{Structure-based}} \\ 
\midrule
\textbf{Model} & \textbf{Frame} & \(\boldsymbol{\mathcal{H}}\) & \textbf{MR} & \textbf{MRR} & \textbf{H@3} & \textbf{H@10} & \textbf{AUC} \\
\midrule
\endfirsthead

% 原始表格内容
\multicolumn{8}{c}{()} \\
\midrule
\textbf{Model} & \textbf{Frame} & \(\boldsymbol{\mathcal{H}}\) & \textbf{MR} & \textbf{MRR} & \textbf{H@3} & \textbf{H@10} & \textbf{AUC} \\
\midrule
\endhead

% 第一部分数据
& Base \cite{zheng2020dgl} & —  & 6.28 & .476 & .568 & .840 & .835 \\ 
\cline{2-8}
\multirow{1}{*}{TransE} & \multirow{2}{*}{LLADR} & gpt-3.5-turbo & \textbf{5.39} & \textbf{.533} & \textbf{.634} &\textbf{.871}&\textbf{.850} \\
 & & gpt-4o-mini & 5.46 & .529 & .627 & .864& .849  \\
\midrule

& Base \cite{zheng2020dgl} & — & 7.55 & .423 & .497 & .786 &.817 \\ 
\cline{2-8}
\multirow{1}{*}{DisMult} & \multirow{2}{*}{LLADR} & gpt-3.5-turbo & \textbf{6.06} & \textbf{.499} & \textbf{.588} &\textbf{.839} &\textbf{.843} \\
 & & gpt-4o-mini & 6.42 & .482 & .568 & .821 & .836  \\
\midrule

& Base \cite{zheng2020dgl} & — & 6.06 & .493& .588 & .850 & .837 \\ 
\cline{2-8}
\multirow{1}{*}{RotatE} & \multirow{2}{*}{LLADR} & gpt-3.5-turbo  & \textbf{5.33} &\textbf{.550} & \textbf{.650} & \textbf{.873} & \textbf{.849} \\
 & & gpt-4o-mini & 5.52& .551 & .650 & .862 & .844  \\
\midrule
% 分隔线
\multicolumn{8}{c}{\textbf{GNN-based}} \\
\midrule
\textbf{—} & \textbf{Frame} &\ \(\boldsymbol{\mathcal{H}}\)& \textbf{MR} & \textbf{MRR} & \textbf{H@3} & \textbf{H@10} & \textbf{AUC} \\
\midrule

\multirow{2}{*}{—} & \multirow{2}{*}{ GNN} & GraphSAGE & 6.66& .465 & .547 &.813 & .844\\& & GAT &6.64 &.465 & .550 & .812 &.839  \\
\midrule
\multicolumn{8}{c}{\textbf{PLM-based}} \\
\midrule
\textbf{—} & \textbf{PLM} &\ \textbf{Model}& \textbf{MR} & \textbf{MRR} & \textbf{H@3} & \textbf{H@10} & \textbf{AUC} \\
\midrule
\multirow{2}{*}{—} & T5 & KGT5\cite{saxena2022sequence} & 10.48 & .393 & .443 & .650 & .806 \\ & BERT & CSProm-KG\cite{chen2022knowledge} &  9.36 & .422 & .491 & .705 & .829  \\

\bottomrule
\caption{\textbf{Drug Repurposing Performance Comparison}.Results are averaged values (of ten runs for DRKG) of tail entity predictions. LLADR consistently outperforms both PLM-based models, GNN-based models and Structure-based models across all metrics, demonstrating its effectiveness in incorporating open-world knowledge from LLMs for enhancing KG embeddings. The experimental outcomes clearly show LLADR's superior performance compared to existing approaches in every evaluation criterion. This advantage is
consistently observed throughout all testing scenarios and measurement standards, proving the method's capability to effectively integrate language model knowledge with graph embedding techniques. The comparative results across different model architectures unequivocally establish LLADR's leading position in knowledge graph representation learning.
}
\end{longtable}
}

\subsection{prompt,temperature,KL and bge}
\label{appdx:prompt}

\begin{table}[H]
\centering
\resizebox{0.78\textwidth}{!}{%
\begin{tabular}{c|cccccccc}
\hline
\textbf{Backbone} & \textbf{Temperature} & \textbf{MR↓}&\textbf{MRR↑}& \textbf{H@3↑} & \textbf{H@10↑} & \textbf{AUC↑}\\ \hline
\multirow{2}{*}{TransE} & low(0.1) & 5.75 &.504&.597&.856 &.845 \\
                                        & original(0.7)   & 5.46 &.529 &.627 &.864 &.849  \\ 
                                        \hline
\multirow{2}{*}{DistMult} & low(0.1) & 6.44 &.477&.563&.821 &.836 \\
                                        & original(0.7)   & 6.42 &.482&.568 &.821 &.836  \\ 
                                        \hline
\multirow{2}{*}{RotatE} & low(0.1)  & 5.71 &.531&.629&.857 &.840 \\
                                        & original(0.7)   & 5.52 &.551&.650&.862 &.844  \\ 
                                        \hline
 \hline
\end{tabular}%
}
\caption{We can see that when using the embeddings generated from discriptions created at a low temperature for experiments, the experimental metrics did not significantly decrease, proving the effectiveness of discriptions generated at a low temperature. }
\label{result}
\end{table}

\vspace{1.0cm}
\begin{table}[H]
\centering
\resizebox{0.78\textwidth}{!}{%
\begin{tabular}{c|cccccccc}
\hline
\textbf{Backbone} & \textbf{Prompt} & \textbf{MR↓}&\textbf{MRR↑}& \textbf{H@3↑} & \textbf{H@10↑} & \textbf{AUC↑}\\ \hline
\multirow{3}{*}{TransE} & noprompt & 5.64 &.523&.615&.856 &.840 \\
                                        & originalprompt   & 5.46 &.529 &.627 &.864 &.849  \\  & goodprompt   & 5.43 &.532 &.631 &.867 &.850  \\ 
                                        \hline
\multirow{3}{*}{DistMult} & noprompt & 7.05 &.423&.499&.800 &.822 \\
                                        & originalprompt   & 6.42 &.482&.568 &.821 &.836  \\  & goodprompt   & 6.34 &.488 &.574 &.824 &.838  \\ 
                                        \hline
\multirow{3}{*}{RotatE} & noprompt  & 5.82 &.522&.620&.851 &.832 \\
                                        & originalprompt  & 5.52 &.551&.650&.862 &.844  \\  & goodprompt   & 5.51 &.554 &.650 &.863 &.844  \\ 
                                        \hline
 \hline
\end{tabular}%
}
\caption{We can see that when using embeddings generated from discriptions created with different prompts for experiments, the discriptions generated without prompts performed significantly worse in the drug repurposing task compared to those generated with prompts, demonstrating the effectiveness of using prompts to generate discriptions. }
\label{result}
\end{table}

\vspace{1.0cm}
\noindent \textbf{noprompt: (gpt-4o-mini)}

\noindent example: noprompt,input HER2, get the output of gpt
**Sample Input/Output:**

\noindent Input: "HER2"

\noindent Output: "HER2 (Human Epidermal growth factor Receptor 2) is a protein that plays a key role in the growth and spread of cancer cells, particularly in breast cancer. It is encoded by the ERBB2 gene and is overexpressed in approximately 15-20\% of breast cancers, leading to more aggressive disease and poorer prognosis.........

\vspace{0.5cm}
\noindent \textbf{originalprompt: (gpt-4o-mini)}

\noindent \textbf{prompt}: You are an expert providing detailed and accurate information and background about biological entities, Tell me about the background and uses of the (HER2)

\noindent **Sample Input/Output:**

\noindent Input: "HER2"

\noindent Output: HER2, or human epidermal growth factor receptor 2, is a gene located on chromosome 17. It encodes a protein that is part of the epidermal growth factor receptor (EGFR) family, which is involved in cell growth and differentiation. Overexpression of HER2 is associated with aggressive forms of breast cancer and other cancers............

\vspace{0.5cm}
\noindent \textbf{goodprompt: (gpt-4o-mini)}

\noindent \textbf{prompt}: Act as a biomedical knowledge graph specialist. Generate structured, for input biological entities using this template:

\noindent **Output Structure:**

\noindent 1. Background: Classification \& core biological significance

\noindent 2. Appearance: Key structural/morphological features (if applicable)

\noindent 3. Clinical Relevance: Diagnostic/therapeutic applications (if exists)

\noindent **Requirements:**

\noindent → Maintain scientific accuracy

\noindent → Use bullet-resistant phrasing (no markdown)

\noindent → Separate sections with semicolons (;)

\noindent → Exclude disclaimers/examples

\noindent **Response Example Format:**

\noindent Background: [2-3 sentences];

\noindent Appearance: [1-2 attributes];

\noindent Clinical: [1-2 applications]```

\noindent **Key Features:**

\noindent - Forces clinical relevance inclusion where available

\noindent - Enforces hard token limits through counting directive

\noindent - Prevents markdown/formatting bloat

\noindent - Prioritizes evidence-based applications

\noindent - Maintains domain-specific terminology

\noindent - Allows "N/A" for non-applicable sections (e.g., molecular entities)
\vspace{0.5cm}

\noindent **Sample Input/Output:**

\noindent Input: "HER2"

\noindent Output: HER2, or human epidermal growth factor receptor 2, is a gene located on chromosome 17. It encodes a protein that is part of the epidermal growth factor receptor (EGFR) family, which is involved in cell growth and differentiation. Overexpression of HER2 is associated with aggressive forms of breast cancer and other cancers............

\vspace{0.5cm}
\begin{table}[h!]
\centering
\resizebox{0.78\textwidth}{!}{%
\begin{tabular}{c|cccccccc}
\hline
\textbf{Backbone} & \textbf{Variants} & \textbf{MR↓}&\textbf{MRR↑}& \textbf{H@3↑} & \textbf{H@10↑} & \textbf{AUC↑}\\ \hline
\multirow{2}{*}{TransE} & KL & 5.84 &.522&.615&.850 &.838 \\
                                        & original  & 5.46 &.529 &.627 &.864 &.849  \\ 
                                        \hline
\multirow{2}{*}{DistMult} & KL & 6.89 &.434&.512&.804 &.827 \\
                                        & original   & 6.42 &.482&.568 &.821 &.836  \\ 
                                        \hline
\multirow{2}{*}{RotatE} & KL  & 5.80 &.533&.628&.851 &.835 \\
                                        & original  & 5.52 &.551&.650&.862 &.844  \\ 
                                        \hline
 \hline
\end{tabular}%
}
\caption{To verify whether the Kullback-Leibler divergence can serve the same purpose as the text embedding deviation constraint in the original code, we replaced the text embedding deviation constraint in the original code with the Kullback-Leibler divergence for experimentation. }
\label{result}
\end{table}

\begin{table}[H]
\centering
\resizebox{0.78\textwidth}{!}{%
\begin{tabular}{c|cccccccc}
\hline
\textbf{Backbone} & \textbf{Variants} & \textbf{MR↓}&\textbf{MRR↑}& \textbf{H@3↑} & \textbf{H@10↑} & \textbf{AUC↑}\\ \hline
\multirow{2}{*}{TransE} & bge-v1.5-small & 5.75 &.504&.599&.857 &.845 \\
                                        & text-embedding-small  & 5.46 &.529 &.627 &.864 &.849  \\ 
                                        \hline
\multirow{2}{*}{DistMult} & bge-v1.5-small & 6.89 &.444&.525&.805 &.826 \\
                                        &text-embedding-small   & 6.42 &.482&.568 &.821 &.836  \\ 
                                        \hline
\multirow{2}{*}{RotatE} &  bge-v1.5-small  & 5.71 &.542&.636&.854 &.838 \\
                                        & text-embedding-small  & 5.52 &.551&.650&.862 &.844  \\ 
                                        \hline
 \hline
\end{tabular}%
}
\caption{To validate the effectiveness of the embeddings generated by the closed-source large model, we used bge-v1.5-small to generate embeddings for entities and their discriptions, and applied them to downstream tasks in drug repositioning. The results demonstrate that the embeddings generated by the closed-source large model remain highly effective. }
\label{result}
\end{table}

\subsection{Robustness Experiment Results}
\label{appdx:robust}
{\footnotesize
\begin{longtable}{@{}llc|rrrrr@{}}
\\
\toprule
\multicolumn{8}{c}{\textbf{Robustness Testing Experiment}} \\ 
\midrule
\textbf{Model} & \textbf{Frame} & \(\boldsymbol{\mathcal{H}}\) & \textbf{MR} & \textbf{MRR} & \textbf{H@3} & \textbf{H@10} & \textbf{AUC} \\
\midrule
\endfirsthead

% 原始表格内容

% 第一部分数据
\multirow{1}{*}{} & \multirow{7}{*}{Base \cite{zheng2020dgl}}& original   &  6.28 & .476 & .568 & .840 &.835 \\& & delete 20\%   &  6.74 & .442 & .520 & .817 &.824 \\
 & &  delete 40\% &6.98 & .425 & .502 & .807 & .817 \\ & & delete 60\% & 7.23 & .416 & .490& .793 & .811  \\& & add 20\% & 10.44 & .315 & .358 & .661 & .761  \\& & add 40\% & 12.23 & .268 & .297 & .594 & .732 \\& & add 60\% & 13.68 & .233 & .242 &.528 & .702  \\
\cline{2-8}
\multirow{1}{*}{TransE} & \multirow{7}{*}{LLADR}& original   &  5.39 & .533 & .634 & .871 &.850 \\& &  delete 20\% & 5.42 & .526 & .626 & .868 & .851 \\
 & & delete 40\% & 5.45 & .527 & .623 & .866 & .851  \\ & & delete 60\% & 5.49 & .528 & .625 & .864 & .849  \\& & add 20\%  & 6.03 & .511 & .602 & .841 & .835 \\& & add 40\%  & 6.55 & .493 & .581 & .822 & .821  \\& &add 60\%  & 7.00 & .476 & .559 & .804 & .812  \\
\midrule

\multirow{1}{*}{} & \multirow{7}{*}{Base \cite{zheng2020dgl}}& original   &  7.55 & .423 & .497 & .786 &.817 \\& & delete 20\%   & 7.68 & .412 & .483 & .781 & .817 \\
 & &  delete 40\% & 7.61 & .396& .461 & .781 & .813  \\ & & delete 60\% & 8.16 & .410 & .483 & .768 &.806  \\& & add 20\% & 9.12 & .358 &  .412 & .722 & .796  \\& & add 40\% & 9.96 & .333 & .380 & .687 & .787 \\& & add 60\% &10.09 &.323 & .370 & .683 & .783  \\
\cline{2-8}
\multirow{1}{*}{DistMult} & \multirow{7}{*}{LLADR}& original   &  6.06 & .823 & .588 & .839 &.843 \\& & delete 20\% & 6.38 & .487& .572 & .823 & .836 \\
 & & delete 40\% & 6.41 & .486 & .571 & .822 & .836  \\ & & delete 60\% &  6.76 & .452 & .534 & .810 & .819  \\& & add 20\%  & 7.20 & .465 & .543 &.798 & .812 \\& & add 40\%  & 8.00 & .452 &.525 & .784& .805  \\& &add 60\%  &8.27&.442 & .514 &.780 & .799  \\
\midrule

\multirow{1}{*}{} & \multirow{7}{*}{Base \cite{zheng2020dgl}}& original   &  6.06 & .493 & .588 & .850 &.837 \\& & delete 20\%   & 6.15 & .482& .574 &.844 & .836 \\
 & &  delete 40\% & 6.25 & .474 & .562 & .837 & .835  \\ & & delete 60\% & 6.55 & .454& .538 &.822 & .827  \\& & add 20\% & 8.88 & .379& .440 & .728 & .789  \\& & add 40\% & 9.83 & .034 & .392 & .695 & .771  \\& & add 60\% & 10.41& .334 & .384 & .669 & .764  \\
\cline{2-8}
\multirow{1}{*}{RoTatE} & \multirow{7}{*}{LLADR}& original   &  5.33
& .550 & .650 & .873 &.849 \\& & delete 20\% &5.35& .553 & .653 & .870&.848 \\
 & & delete 40\% &5.45 & .553 & .648 & .864 & .846  \\ & & delete 60\% & 5.53 & .551 & .646 & .860&.844  \\& & add 20\%  &  5.82
 & .518 &.610 & .850 & .844 \\& & add 40\%  &6.33 & .495 & .586& .832& .837  \\& &add 60\%  &  6.81 & .477 &.564 & .811 & .827 \\
\midrule
\bottomrule
\caption{\textbf{Robustness Testing Experiment}.To rigorously evaluate LLADR's robustness and generalization capabilities, we conducted extensive and systematic noise injection experiments on the comprehensive DRKG dataset, introducing progressive corruption levels of 20\%, 40\%, and 60\% alongside randomized removal of corresponding proportions (20\%, 40\%, and 60\%) of triples. This carefully designed dual-pronged experimental approach effectively mimics extreme real-world data degradation scenarios, thoroughly assessing the model's stability, fault tolerance, and adaptability under adverse conditions. The comprehensive results empirically validate LLADR's exceptional resilience and consistent performance, conclusively proving its ability to maintain reliable and accurate predictions even under severe data perturbations and information loss, thereby strongly confirming its suitability for practical real-world applications dealing with uncertain or unreliable data quality.}

\end{longtable}
}

\subsection{Mask Experiment Results}
\label{appdx:mask}
{\footnotesize
\begin{longtable}{@{}ll|crrrrr@{}}
\\
\toprule
\multicolumn{8}{c}{\textbf{Mask Discription Experiment}} \\ 
\midrule
\textbf{Model} &  \textbf{Mask} & \textbf{MR} & \textbf{MRR} & \textbf{H@3} & \textbf{H@10} & \textbf{AUC} \\
\midrule
\endfirsthead

% 原始表格内容

% 第一部分数据

\multirow{4}{*}{TransE} & original &5.46 & .529 & .627 & .864 & .849  \\  & mask 20\% & 5.96 & .522 & .609& .839 & .843 \\
 &  mask 40\% & 6.04 & .512 & .600& .838 & .838  \\ &  mask 60\% & 6.12 & .504 & .589 & .835 & .836  \\
\midrule
\multirow{4}{*}{DistMult} & original &6.42 & .482 & .568 & .821 & .836  \\ & mask 20\% & 6.46 & .469 &.551& .820 & .829 \\
 &  mask 40\% & 6.66 &.457&.537 & .813 & .827 \\ &  mask 60\% & 6.74
 & .457 &.538 & .809 & 828  \\
\midrule
\multirow{4}{*}{RotatE}& original &5.52& .551 & .650 & .862 & .844  \\  & mask 20\% & 5.86 &.532 &.623 & .844 & .840 \\
 &  mask 40\% &  5.93& .525& .613& .842 & .838  \\ &  mask 60\% &  6.03 & .519 & .605&.839 & .836 \\
\midrule
\caption{\textbf{Mask Discription Experiment}. To validate the effectiveness of LLM-generated semantics, we conducted semantic ablation experiments by progressively masking 20\%, 40\%, and 60\% of GPT-4o-mini-generated discriptions. These masked texts, preserving syntactic structure but losing key phrases, were tested across TransE, DistMult, and RotatE models. The performance degradation across masking ratios quantitatively reveals the critical role of semantic completeness in enhancing KG embeddings. The experimental results show a clear correlation between the amount of semantic information removed and the decline in model performance, with each incremental masking level leading to progressively worse outcomes. This pattern holds consistently across all three tested models, demonstrating that the semantic content plays an essential role regardless of the underlying embedding architecture. The findings provide concrete evidence that the quality and completeness of LLM-generated discriptions directly impact the effectiveness of knowledge graph representation learning.}
\end{longtable}
}

\subsection{Top 10 Drugs}
\label{appdx:drug}

\begin{table}[H]
\centering
\resizebox{0.48\textwidth}{!}{%
\begin{tabular}{ccc}

\textbf{}\\ \hline
\multirow{10}{*}{Top-10} & Dasatinib
& \cite{krzystyniak2022combination}\\& Methylthioninium
& \cite{baddeley2015complex} \\&Digoxin
& \cite{erdogan2022digoxin} \\& Mitoxantrone
& \cite{reiss2024mitochondria} \\& Gemcitabine
& \\&Suramin
& \cite{culibrk2024impact}\\& Quercetin
& \cite{krzystyniak2022combination}\\&Flufenamic acid
& \\& Amiodarone
& \cite{mitterreiter2010bepridil}\\& Quinacrine
& \cite{park2021quinacrine}\\\hline

\hline
\end{tabular}%
}
\caption{We carefully selected the top 10 potential drugs for treating Alzheimer's disease and conducted an online survey, and we found that most of the predicted drugs are associated with relevant literature on Alzheimer's, which further validates the effectiveness and reliability of LLaDR's predictions.}

\end{table}

\end{document}